\documentclass[twoside,twocolumn,9pt]{article}
\usepackage{extsizes}
\usepackage[super,sort&compress,comma]{natbib} 
\usepackage[version=3]{mhchem}
\usepackage[left=1.5cm, right=1.5cm, top=1.785cm, bottom=2.0cm]{geometry}
\usepackage{balance}
\usepackage{widetext}
\usepackage{times,mathptmx}
\usepackage{sectsty}
\usepackage{graphicx} 
\usepackage{lastpage}
\usepackage[format=plain,justification=raggedright,singlelinecheck=false,font={stretch=1.125,small,sf},labelfont=bf,labelsep=space]{caption}
\usepackage{float}
\usepackage{fancyhdr}
\usepackage{fnpos}
\usepackage[english]{babel}
\usepackage{array}
\usepackage{droidsans}
\usepackage{charter}
\usepackage[T1]{fontenc}
\usepackage[usenames,dvipsnames]{xcolor}
\usepackage{setspace}
\usepackage[compact]{titlesec}
\usepackage{hyperref}

\usepackage{epstopdf}

\definecolor{cream}{RGB}{222,217,201}

\begin{document}

\pagestyle{fancy}
\thispagestyle{plain}
\fancypagestyle{plain}{

\fancyhead[C]{\includegraphics[width=18.5cm]{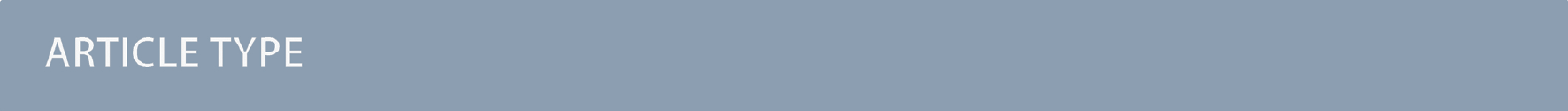}}
\fancyhead[L]{\hspace{0cm}\vspace{1.5cm}\includegraphics[height=30pt]{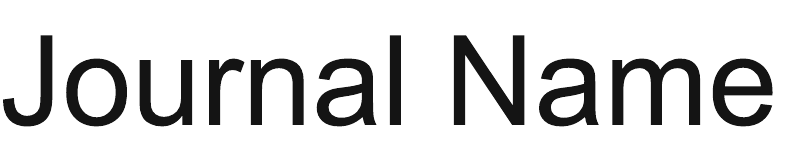}}
\fancyhead[R]{\hspace{0cm}\vspace{1.7cm}\includegraphics[height=55pt]{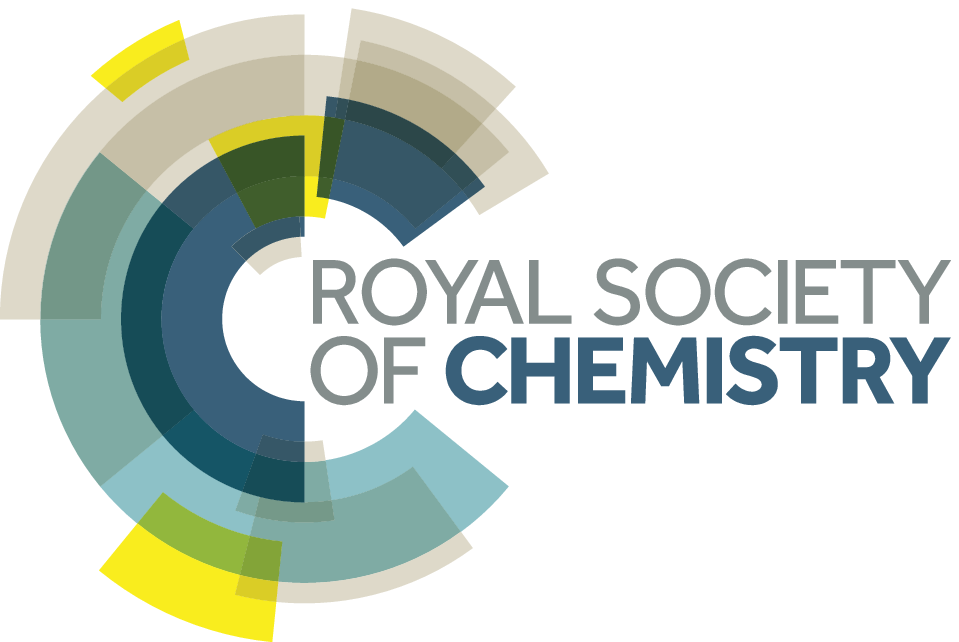}}
\renewcommand{\headrulewidth}{0pt}
}

\makeFNbottom
\makeatletter
\renewcommand\LARGE{\@setfontsize\LARGE{15pt}{17}}
\renewcommand\Large{\@setfontsize\Large{12pt}{14}}
\renewcommand\large{\@setfontsize\large{10pt}{12}}
\renewcommand\footnotesize{\@setfontsize\footnotesize{7pt}{10}}
\makeatother

\renewcommand{\thefootnote}{\fnsymbol{footnote}}
\renewcommand\footnoterule{\vspace*{1pt}%
\color{cream}\hrule width 3.5in height 0.4pt \color{black}\vspace*{5pt}} 
\setcounter{secnumdepth}{5}

\makeatletter 
\renewcommand\@biblabel[1]{#1}            
\renewcommand\@makefntext[1]%
{\noindent\makebox[0pt][r]{\@thefnmark\,}#1}
\makeatother 
\renewcommand{\figurename}{\small{Fig.}~}
\sectionfont{\sffamily\Large}
\subsectionfont{\normalsize}
\subsubsectionfont{\bf}
\setstretch{1.125} 
\setlength{\skip\footins}{0.8cm}
\setlength{\footnotesep}{0.25cm}
\setlength{\jot}{10pt}
\titlespacing*{\section}{0pt}{4pt}{4pt}
\titlespacing*{\subsection}{0pt}{15pt}{1pt}

\fancyfoot{}
\fancyfoot[LO,RE]{\vspace{-7.1pt}\includegraphics[height=9pt]{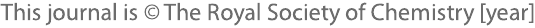}}
\fancyfoot[CO]{\vspace{-7.1pt}\hspace{13.2cm}\includegraphics{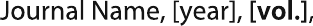}}
\fancyfoot[CE]{\vspace{-7.2pt}\hspace{-14.2cm}\includegraphics{RF}}
\fancyfoot[RO]{\footnotesize{\sffamily{1--\pageref{LastPage} ~\textbar  \hspace{2pt}\thepage}}}
\fancyfoot[LE]{\footnotesize{\sffamily{\thepage~\textbar\hspace{3.45cm} 1--\pageref{LastPage}}}}
\fancyhead{}
\renewcommand{\headrulewidth}{0pt} 
\renewcommand{\footrulewidth}{0pt}
\newcommand\tab[1][0.5cm]{\hspace*{#1}}
\newcommand\norm[1]{\left\lVert#1\right\rVert}
\setlength{\arrayrulewidth}{1pt}
\setlength{\columnsep}{6.5mm}
\setlength\bibsep{1pt}

\makeatletter 
\newlength{\figrulesep} 
\setlength{\figrulesep}{0.5\textfloatsep} 

\newcommand{\topfigrule}{\vspace*{-1pt}%
\noindent{\color{cream}\rule[-\figrulesep]{\columnwidth}{1.5pt}} }

\newcommand{\botfigrule}{\vspace*{-2pt}%
\noindent{\color{cream}\rule[\figrulesep]{\columnwidth}{1.5pt}} }

\newcommand{\dblfigrule}{\vspace*{-1pt}%
\noindent{\color{cream}\rule[-\figrulesep]{\textwidth}{1.5pt}} }

\makeatother

\twocolumn[
  \begin{@twocolumnfalse}
\vspace{3cm}
\sffamily
\begin{tabular}{m{4.5cm} p{13.5cm} }

\includegraphics{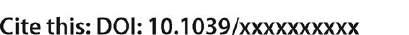} & \noindent\LARGE{\textbf{Deeply learning molecular structure-property relationships using attention- and gate-augmented graph convolutional network}} \\
\vspace{0.3cm} & \vspace{0.3cm} \\

 & \noindent\large{Seongok Ryu,\textit{$^{a}$} Jaechang Lim,\textit{$^{a}$} Seung Hwan  Hong,\textit{$^{a}$} and Woo Youn Kim$^{\ast}$\textit{$^{a,b}$}} \\

\includegraphics{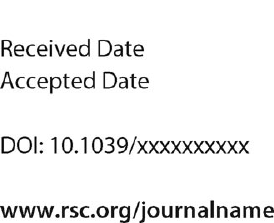} & \noindent\normalsize{
Molecular structure-property relationships are key to molecular engineering for materials and drug discovery. The rise of deep learning offers a new viable solution to elucidate the structure-property relationships directly from chemical data. Here we show that the performance of graph convolutional networks (GCNs) for the prediction of molecular properties can be improved by incorporating attention and gate mechanisms. The attention mechanism enables a GCN to identify atoms in different environments. The gated skip-connection further improves the GCN by updating feature maps at an appropriate rate. We demonstrate that the resulting attention- and gate-augmented GCN could extract better structural features related to a target molecular property such as solubility, polarity, synthetic accessibility and photovoltaic efficiency compared to the vanilla GCN. More interestingly, it identified two distinct parts of molecules as essential structural features for high photovoltaic efficiency, and each of them coincided with the areas of donor and acceptor orbitals for charge-transfer excitations, respectively. As a result, the new model could accurately predict molecular properties and place molecules with similar properties close to each other in a well-trained latent space, which is critical for successful molecular engineering.
} \\

\end{tabular}

 \end{@twocolumnfalse} \vspace{0.6cm}
]

\renewcommand*\rmdefault{bch}\normalfont\upshape
\rmfamily
\section*{}
\vspace{-1cm}


\footnotetext{\textit{$^{a}$~Department of Chemistry, KAIST, 291 Daehak-ro, Yuseong-gu, Daejeon 34141, Republic of Korea}}
\footnotetext{\textit{$^{b}$~KI for Artificial Intelligence, KAIST, 291 Daehak-ro, Yuseong-gu, Daejeon 34141, Republic of Korea }}
\footnotetext{$^{\ast}$~Correspondence and requests for materials should be addressed to W.Y.K. (email : wooyoun@kaist.ac.kr).}
\footnotetext{Electronic Supplementary Information (ESI) available: See DOI: 10.1039/b000000x/}



\section{Introduction}

Elucidating molecular structure-property relationships plays a pivotal role for successful molecular design. However, the structure-property relationships are usually unknown, so molecular engineering based on heuristic rules inevitably goes through multiple trial and error. As a more systematic approach, a computer-aided molecular design has attracted great attention, especially in drug and materials discovery.\cite{mccammon1987computer, todd2005computer} Promising molecules with desired properties are first selected through the high-throughput virtual screening of a large library using computational chemistry before experiments.\cite{shoichet2004virtual, curtarolo2013high, pyzer2015high} During the procedure, high computational costs are necessary to obtain reliable results. Practical approximations adopted for efficient screening provoke undesirable errors, making the screening result less reliable.

The rise of deep learning (DL)\cite{lecun2015deep}  techniques is expected to open a new paradigm for efficient molecular design. Unlike traditional computational methods based on physical principles, the DL can find out structure-property relationships directly from chemical data and apply it in new ways to molecular design. For example, supervised learning methods have been widely used to learn and predict molecular energetics,\cite{rupp2012fast, gilmer2017neural, schutt2017quantum, schutt2017schnet, smith2017ani, faber2017prediction} toxicity,\cite{duvenaud2015convolutional, kearnes2016molecular, wu2018moleculenet, mayr2016deeptox} and drug efficacy\cite{gomes2017atomic, ozturk2018deepdta, jimenez2018k}. A class of unsupervised learning in particular with novel generative models has been utilized for \textit{de novo} molecular design. \cite{gomez2018automatic, kusner2017grammar, segler2017generating, gupta2018generative, jaques2016sequence, kang2018conditional, lim2018molecular, müller2018recurrent, li2018learning, you2018graphrnn, jin2018junction, simonovsky2018graphvae, li2018multi, olivecrona2017molecular, guimaraes2017objective, de2018molgan, you2018graph} Reinforcement learning techniques facilitate designing drugs and planning synthetic routes.\cite{guimaraes2017objective, de2018molgan, you2018graph, chen2018using, zhou2017optimizing, segler2018planning, wei2016neural} As such, the DL, although at an early stage, is rapidly spreading to various chemical fields as a complement of traditional computational chemistry.

The key to success of DL in chemistry is elucidating correct structure-property relationships from existing data. That is equivalent to constructing a DL model to best approximate a function $f$ in $Y = f(X)$, where $Y$ and $X$ denote molecular properties and structures, respectively.   In principle, it is possible to construct the model if a vast amount of high quality data are available because DL is known as a universal approximation kernel.\cite{lin2017does} However, in practice, lack of chemical data limits its wide applications. Thus, it is essential to develop a high performance DL model specialized for chemical problems, as can be seen from the great success of convolutional neural networks (CNNs) in vision recognition,\cite{krizhevsky2012imagenet, kim2014convolutional} and recurrent neural networks (RNNs) in natural language processing.\cite{mikolov2013efficient, cho2014learning, bowman2015generating} A high performance DL model is able to extract important structural features determining a target property from limited data. In chemistry, both CNN and RNN have been used to process molecules represented with SMILES\cite{gomez2018automatic, kusner2017grammar, segler2017generating, gupta2018generative, jaques2016sequence, kang2018conditional, lim2018molecular} and molecular fingerprints\cite{duvenaud2015convolutional, kearnes2016molecular, wu2018moleculenet}, since those models are readily available. However, SMILES and fingerprints are too simple to deliver the topological information of molecular structures, and leads to relatively a low learning accuracy. Within DL models, \citeauthor{schutt2017quantum} proposed namely the deep tensor neural network model and achieved a high accuracy for molecular energetics by exploiting 3D molecular structures.\cite{schutt2017quantum} Unfortunately, most chemical data provides simplified molecular structures such as SMILES, fingerprints, and molecular graphs, with which calculations of 3D molecular structures are very demanding. 

In this aspect, a molecular graph representation would be the best compromise; it describes atoms and bonds in a molecule as nodes and edges, respectively. Molecular graphs intuitively and concisely express molecules with 2D topological information. Hence, they are widely adopted in chemical education as well as chemical informatics. Indeed, there have been efforts to develop DL models based on molecular graphs. Graph convolutional network (GCN), as an extension of the CNN, was proposed to deal with graph structures.\cite{kipf2016semi, defferrard2016convolutional} The GCN benefits from the advantage of the CNN architecture; it performs with a high accuracy but a relatively low computational cost by utilizing fewer parameters compared to a fully connected multi-layer perceptron (MLP) model. It can also identify important atom features that determine molecular properties by analyzing relations between neighboring atoms.\cite{duvenaud2015convolutional} Weave model, a variant of the GCN, considers not only atom features but also bond features.\cite{kearnes2016molecular}

Despite the aforementioned advantages, we suspect that the GCN is still missing an important structural feature to learn better structure-property relationships. Molecules are not just a simple collection of atoms. Same atoms can often cause different molecular properties depending on their local chemical environments. For instance, carbon atoms of aromatic rings, aliphatic chains, and carbonyl groups have different characters owing to their different chemical environments. Chemists can identify functional groups related to molecular properties. Polar and nonpolar groups are examples of such molecular polarity and solubility. Therefore, it is critical to correctly identify molecular substructures, which determine a target property, to learn more accurate structure-property relationships. However, previous models apply identical convolution weights to all atoms and bonds. In other words, they treat all atoms and bonds with equal importance regardless of their chemical environments. 

To improve performance of the GCN, one possible approach is to add adaptive attention weights depending on chemical environments to graph convolutions. The resulting neural network optimizes both convolution and attention weights in a learning process of the structure-property relationships. The so-called graph attention network was originally developed for a network analysis in computer science.\cite{velickovic2017graph} In chemistry, \citeauthor{shang2018edge} first adopted the attention mechanism for prediction of molecular properties.\cite{shang2018edge} It should be noted that \citeauthor{shang2018edge} described molecules as an assembly of chemical bonds. Thus, they applied the attention mechanism to chemical bonds and then used the same bond features across all molecules. For instance, all C=O bonds share the same bond features. However, C=O of a carboxyl group is chemically different from that of an ester group (e.g., bond length and strength), which means that bond features must depend on chemical environments as well. Therefore, chemical bonds would not be appropriate building blocks for molecules.

Another way to improve the GCN is to adjust the update rate of node states at each convolution layer. As propagating through convolution layers in the forward direction, a node state is updated by gradually considering from the nearest neighboring nodes to those at far distances. The vanilla flavor of GCN has no way to determine the best updating rate during the propagation. In fact, \citeauthor{kipf2016semi} reported that the GCN was not trained properly with more than 7 convolution layers.\cite{kipf2016semi} On the other hand,  the gated graph neural network (GGNN)\cite{li2015gated} does not show such a problem because it can find out an appropriate update rate via a gate mechanism implied in a gated recurrent unit cell (see Fig. 3(a) in this work).  

In this regard, we propose to incorporate the attention and gate mechanisms in the GCN to more deeply learn molecular structure-property relationships. The attention mechanism can differentiate atoms in different chemical environments by considering an interaction of each atom with neighbors. Its pairwise interactions are similar to the atom pair concept in Shang's model. The main difference from Shang's model is that our model may have different interactions even for identical atom pairs if they are in different chemical environments and so have different atom features. Therefore, our model more flexibly elucidates the structure-property relationships. As a cost for the flexibility, attention weights for pairwise interactions should be trained for all atom pairs independently. After updating atom states with the attention mechanism, the updated and previous states are appropriately combined by a gated skip-connection. This mitigates not only the vanishing gradient problem, but also the accuracy reduction issue caused by stacked graph convolution layers.

Here we focus on the role of the attention and gate mechanisms in clarifying the structure-property relationships. We specifically show that the augmented GCN is able to identify important molecular substructures which are directly related to target properties. In addition, it can chemically rationalize the important structural features by mapping them on molecular graphs. This is important because scientific interpretation of a result is often more valuable than the result itself. Not surprisingly, the augmented GCN can recognize polar and nonpolar functional groups as important structural features for molecular solubility and polarity. In addition, we show that the augmented GCN can distinguish two separated molecular regions related to charge-transfer excitations for highly efficient photovoltaic molecules without any electronic structure information. We confirmed that the two regions in fact coincide with the donor and acceptor orbital regions, respectively. Apparently, it is not a trivial task even to experts without information on the electronic structure of molecules.

\section{Theoretical backgrounds}

\subsection{Graph representation of molecules}

Fig. 1 shows an example of molecular graph inputs in this study. Each molecular graph ($G$) consists of initial node features, $X_i$, and an adjacency matrix, $A$. The node features correspond to atom descriptors including atom type, number of attached hydrogens, number of valencies, and aromaticity indicator. We represent all the descriptors with an one-hot encoded vector. An adjacency matrix represents only the connectivity between atom pairs. In other words, it does not include explicit edge descriptors. For example, single and double bonds are equally represented because they can be deduced from the corresponding atom descriptors. We obtained the atom descriptors and the adjacency matrices of molecules from the open-source python toolkit, RDKit\cite{landrum2006rdkit}.

\begin{figure}[t]
\centering
  \includegraphics[width=8.5cm]{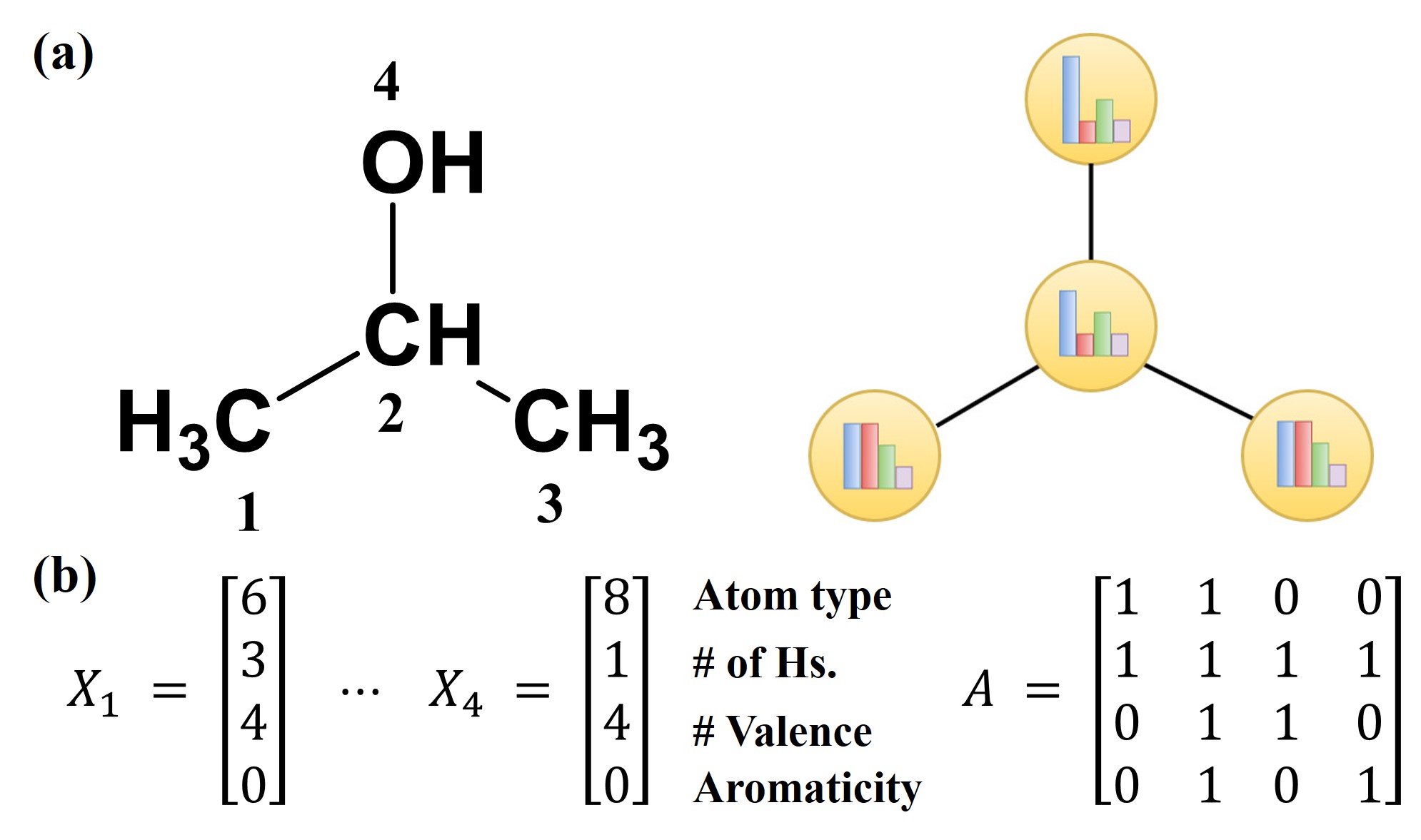}
  \caption{(a) Graph representation of 2-propanol, $G(A, X)$. The colored columns in each node represent atom descriptors, $X_i$. (b) The $i$-th atom descriptors $X_i$ contains initial atom features (atom type, number of hydrogens attached, number of valence electrons, and aromaticity) and the adjacency matrix $A$ represents the connectivity between atom pairs including self-connections.}
  \label{fgr:fig1}
\end{figure}

\begin{figure*} 
 \centering
 \includegraphics[width=18cm]{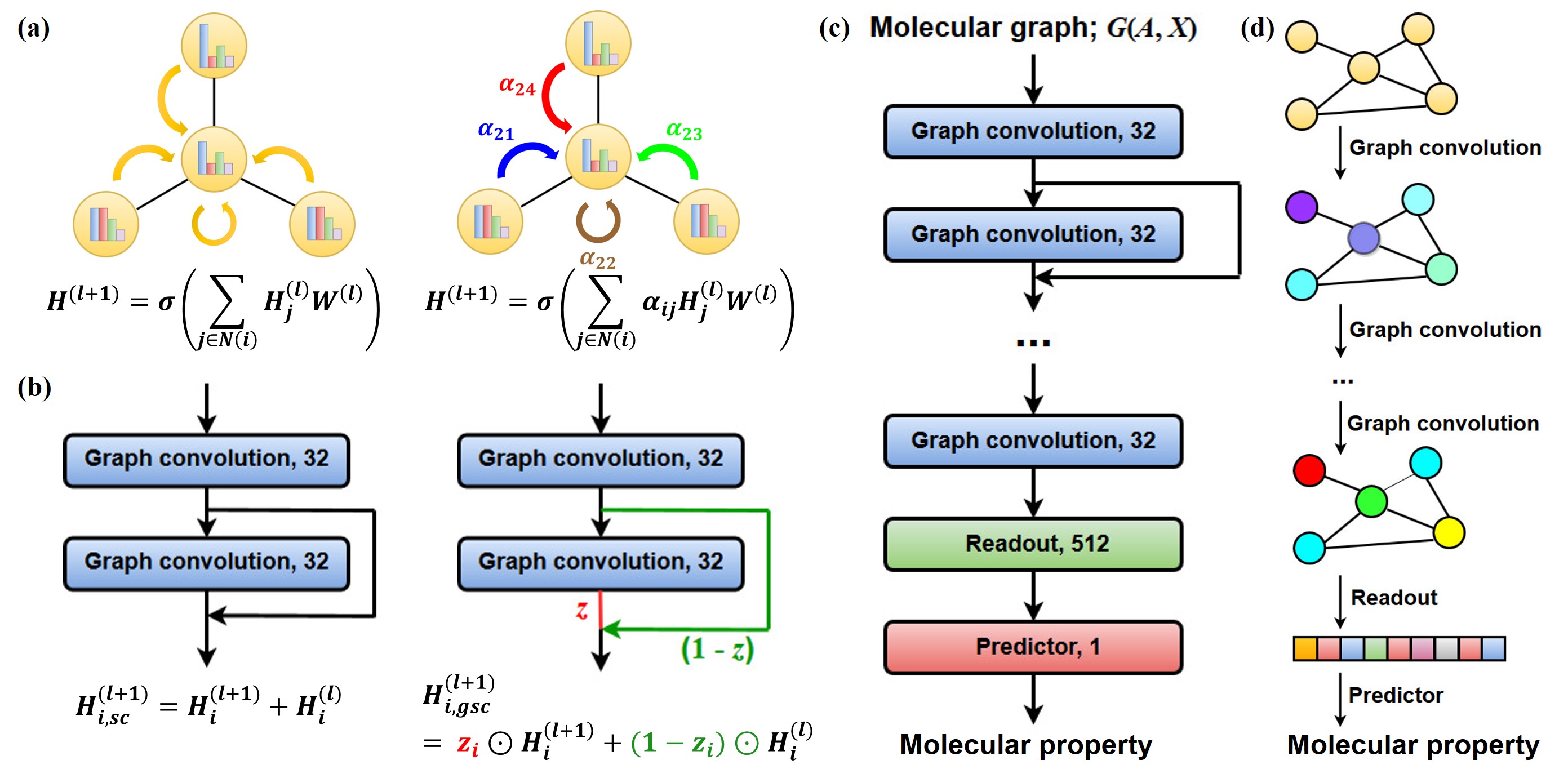}
 \caption{Schematic description of (a) a graph convolution (\textit{left}) and its attention-augmented version (\textit{right}). Arrows depict the transfer of neighboring nodes' information, and different colors of the arrows reflect importance of each neighboring node. (b) Skip-connection (sc, \textit{left}) and gated skip-connection (gsc, \textit{right}). An update rate z is obtained by eq. \eqref{eqn:eq11}. (c) Architecture of graph convolution networks in this work. It is composed of graph convolution layers, a readout, and a predictor. (d) Overall procedure from updating atom features to obtaining a target property which is processed by the neural network presented in (c).}
 \label{fgr:fig2}
\end{figure*}

\subsection{Graph convolution for updating atom states}

Various node embedding methods exist for updating node states in a graph structure. The most general form introduced by \citeauthor{gilmer2017neural} adopts a message passing framework given by
\begin{equation}\label{eqn:eq1}
  H_i^{(l+1)} = U(H_i^{(l)}, m_i^{(l+1)}),
\end{equation}
which describes that the $i$-th node state is updated as a function of the previous node state, $H_i^{(l)}$, and a message state containing pair-wise interaction terms with its neighbors, $m_i^{(1+1)}$.\cite{gilmer2017neural} The GCN is the simplest version of namely the message passing neural network. GCN updates each node state of the $l$-th layer, $H_i^{(l+1)}$, as follows.\cite{kipf2016semi}
\begin{equation} \label{eqn:eq2}
	H^{(l+1)} = \sigma(AH^{(l)}W^{(l)}),
\end{equation}
where $\sigma(\cdot)$, $A$, and $W^{(l)}$ denote an activation function, an adjacency matrix (Fig. 1(a)), and convolution weights of the $l$-th layer, respectively. Suppose that atom 2 has three adjacent atoms 1, 3 and 4, as shown in Fig. 2(a). In this case, equation (2) can be rewritten as
\begin{equation}\label{eqn:eq3}
H_{2} ^{(l+1)} = \sigma(H_{2}^{(l)}W^{(l)}+H_{1}^{(l)}W^{(l)}+H_{3}^{(l)}W^{(l)}+H_{4}^{(l)}W^{(l)}),
\end{equation}
which means that the graph convolution adds up all atom features with same convolution weights and then puts the result through the activation function to output the updated node feature. A single graph convolution updates atom features only from its adjacent atoms as depicted in Fig. 2. The GCN is cost-efficient for it requires a small number of weight parameters. However, it lacks important terms; for example, i) an explicit representation of importance of the relationship between a central atom and its neighbors and ii) an update rate of the previous atom state. It is expected that the node updating can be improved by incorporating the two missing terms in the GCN, which will be discussed in what follows.

\subsection{Attention and gated skip-connection}
One of the crucial keys to the success of natural language processing is attention mechanism. The attention mechanism, for example in machine translation, finds the relationship between words and leverages neural network's ability to find the best word to be translated.\cite{bahdanau2016end} The attention mechanism can be applied to the GCN to capture the relationship between adjacent atoms according to relative importance, which is related to the first missing term above. Then, the atom state updating in eq. \eqref{eqn:eq3} is rewritten as a linear combination of neighboring atom states with attention coefficients (Fig. 2(a) - \textit{right}).
\begin{multline} \label{eqn:eq4}
H_{2} ^{(l+1)} = \sigma(\alpha_{22}^{(l)}H_{2}^{(l)}W^{(l)}+\alpha_{21}^{(l)}H_{1}^{(l)}W^{(l)} \\
+\alpha_{23}^{(l)}H_{3}^{(l)}W^{(l)}+\alpha_{24}^{(l)}H_{4}^{(l)}W^{(l)}),
\end{multline}
where $\alpha_{ij}^{(l)}$ denotes an attention coefficient which measures the importance of the $j$-th node in updating the $i$-th state of the $l$-th hidden layer. The most general expression of the attention coefficient may be as follows.
\begin{equation} \label{eqn:eq5}
\alpha_{ij}^{(l)} = f(H_i^{(l)}W^{(l)},H_j^{(l)}W^{(l)}) 
\end{equation}
The attention coefficient can be obtained typically by i) a similarity base, ii) concatenating features, and iii) coupling all features.\cite{luong2015effective} For example, the first GCN with the attention mechanism\cite{velickovic2017graph} used a concatenation of node features, as below.
\begin{equation} \label{eqn:eq6}
\alpha_{ij} = \frac{e_{ij}}{\sum_{k \in N(i)} e_{ik}} = \frac{\sigma(MLP[H_iW,H_jW])}{\sum_{k \in N(i)} \sigma(MLP[H_iW,H_kW])},
\end{equation}
where $MLP$ stands for a multi-layer perceptron and $[\cdot, \cdot]$ is a concatenation of two matrices. Since the attention coefficient is obtained from the softmax function, eq. \eqref{eqn:eq6} can be interpreted as the importance rate of each adjacent node. For molecular applications, however, the attention coefficient should be analogous to the interaction strength between an atom pair $(i, j)$, instead of the importance ratio between adjacent nodes, to best predict a target property. Thus, we evaluate the attention coefficient through the coupling between atom pairs as follows:
\begin{equation} \label{eqn:eq7}
\alpha_{ij}^{(l)} = \sigma((H_i^{(l)}W^{(l)})C^{(l)}(H_j^{(l)}W^{(l)})^T),
\end{equation}
where $C^{(l)}$ is a coupling matrix. Note that the coupling matrix may correspond to the dictionaries containing pairwise interactions in Shang's model.  In our case, however, the coupling matrix is determined for every atom pair in a given molecule and thus may be different for same atom pairs in different chemical environment. Also, we use a multi-head attention that utilizes $K$-different channels to describe molecular interactions, as shown in eq. \eqref{eqn:eq8}.
\begin{equation} \label{eqn:eq8}
H_i^{(l+1)} = \sigma( \frac{1}{K} \sum_{k=1}^{K} \sum_{j \in N(i)} \alpha_{ij,k}^{(l)} H_{j}^{(l)} W^{(l)})
\end{equation}
We use $ReLU(x) = max(0,x)$ for the activation functions in equations (2) and (8) and $tanh(x) = \frac{e^x - e^{-x}}{e^x + e^{-x}}$  for equation (7). We use $K=4$ for our implementations. 

To reflect atom features at long distances in a specific central atom, a multiple number of graph convolution is applied. Thus, a stack of graph convolution layers is necessary. As is well known, however, deeply stacked layers have side-effects, such as the vanishing gradient problem. Using skip-connection in each hidden layer is a common solution to remedy the problem, and the same idea can be equally applied to the GCN, as below.
\begin{equation} \label{eqn:eq9}
H_{i,sc}^{(l+1)} = H_i^{(l+1)} + H_i^{(l)}
\end{equation}
Here, we need to take one more fact into account when using the GCN. Performing multiple graph convolutions enables the model to carry over information from distant atoms.  \citeauthor{kipf2016semi} reported effects of the model depth for GCN.\cite{kipf2016semi} Without the skip-connection, they could not train models with more than 7 graph convolution layers. Although using the skip-connection allowed them to achieve deeper models, accuracy of the models was gradually lowered as the number of convolution layers increases. We also observed the similar results (see in Fig. 3(a)).

To avoid the aforementioned problems, which is related to the second missing term, we propose a gated skip-connection. The concept of gate is used in recurrent cell units such as the gated recurrent unit (GRU)\cite{cho2014learning} and the long-short term memory (LSTM)\cite{hochreiter1997long}. When a model updates sequential hidden states, the gate is used to deliver previous information accurately by determining forget and update rates. Inspired from the gate mechanism in the recurrent cell units, we adopt the gate in using the skip-connection.
\begin{equation} \label{eqn:eq10}
H_{i,gsc}^{(l+1)} = z_i \odot H_i^{(l+1)} + (1 - z_i) \odot H_i^{(l)}
\end{equation}
with
\begin{equation} \label{eqn:eq11}
z_i = f(H_i^{(l+1)}, H_i^{(l)}) = \sigma( U_{z,1} H_i^{(l+1)} + U_{z,2} H_i^{(l)} + b_z),
\end{equation}
where $U_{z,1}$, $U_{z,2}$, $b_z$ are trainable parameters, $\odot$ is element-wise matrix multiplication (Hadamard product) and $\sigma(\cdot)$ in eq. \eqref{eqn:eq11} is sigmoid activation, which is used to determine the range of activation from 0 to 1.

We augmented the vanilla GCN with the attention, the gated skip-connection, and both of them, respectively, and evaluated their relative performances for prediction of various molecular properties. Table 1 summarizes the atom state updating methods used in this study.

\begin{table}[h]
\small
  \caption{Atom sate updating methods for GCN, GCN+attention, GCN+gate, and GCN+attention+gate. $\alpha_{ij}$ and $z_i$ are given by eq. \eqref{eqn:eq7} and \eqref{eqn:eq11}, respectively.}
  \label{tbl:example}
  \begin{tabular*}{0.5\textwidth}{@{\extracolsep{\fill}}lll}
    \hline
    Model & Atom state updating, $H_i^{(l+1)}$\\
    \hline
    GCN & $\sigma(\sum_{j \in N(i)} H_j^{(l)}W^{(l)})$\\
    GCN+attention & $\sigma(\frac{1}{K}\sum_{k=1}^{K}\sum_{j \in N(i)} \alpha_{ij}^{(l)} H_j^{(l)}W^{(l)})$\\
    GCN+gate & $z_i \odot \sigma( \sum_{j \in N(i)} H_j^{(l)}W^{(l)}) + (1-z_i) \odot H_i^{(l)}$\\
    GCN+attention+gate & $z_i \odot \sigma(\frac{1}{K}\sum_{k=1}^{K}\sum_{j \in N(i)} \alpha_{ij}^{(l)} H_j^{(l)}W^{(l)}) + (1-z_i) \odot H_i^{(l)}$\\
    \hline
  \end{tabular*}
\end{table}

\subsection{Readout and prediction of molecular property}
It is important to note that the graph structures and related features satisfy permutation invariance. Permutation invariance is trivial in node-wise classifications. However, it must be satisfied when the entire graph corresponds to one label, such as in molecular property prediction. After finalizing atom states update, we obtain a graph feature (molecular feature) by gathering the atom states, as described in Fig. 2(a). A readout function is used to process the graph feature, and the most typical choice is a summation of all atom states processed by a MLP.
\begin{equation} \label{eqn:eq12}
z_G = \sum_{i \in G} MLP(H_i^{(L)})
\end{equation}

The summation over all nodes secures the permutation invariance of graph features. Finally, the predictor outputs the molecular properties obatined from the graph features.
\begin{equation} \label{eqn:eq13}
y_{pred} = MLP(z_G)
\end{equation}

To summarize, our models consist of three parts - i) Graph convolution layers augmented with the attention and/or gated skip-connection methods update atom features, ii) Readout function gathers all atom features and generates graph features with permutation invariance, and  iii) Predictor predicts molecular properties from the graph features. Atom feature and graph feature vectors for further analysis are sampled from the graph convolution and readout layers, respectively. Fig. 2(d) visualizes the overall procedure of updating atom features and obtaining a target property processed by the graph convolutional network depicted in Fig. 2(c).

\section{Results and Discussion}
\subsection{Effects of attention and gated skip-connection on performance of GCN 
}

\begin{figure}[t]
\centering
  \includegraphics[width=8cm]{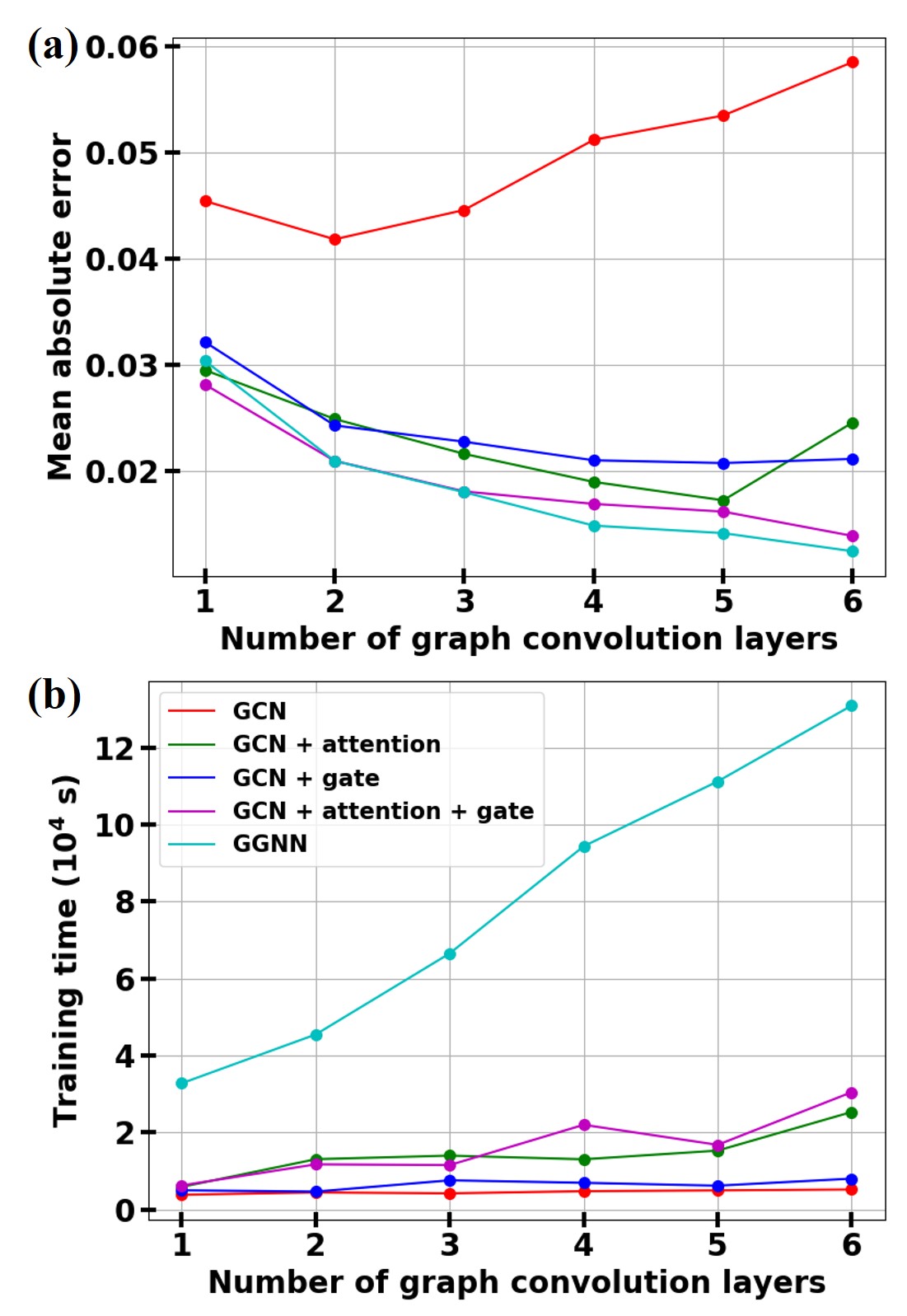}
  \caption{(a) Mean absolute error, and (b) total training time for logP prediction with respect to the number of graph convolution layers.}
  \label{fgr:fig3}
\end{figure}

\begin{figure}[h!]
 \centering
 \includegraphics[width=8cm]{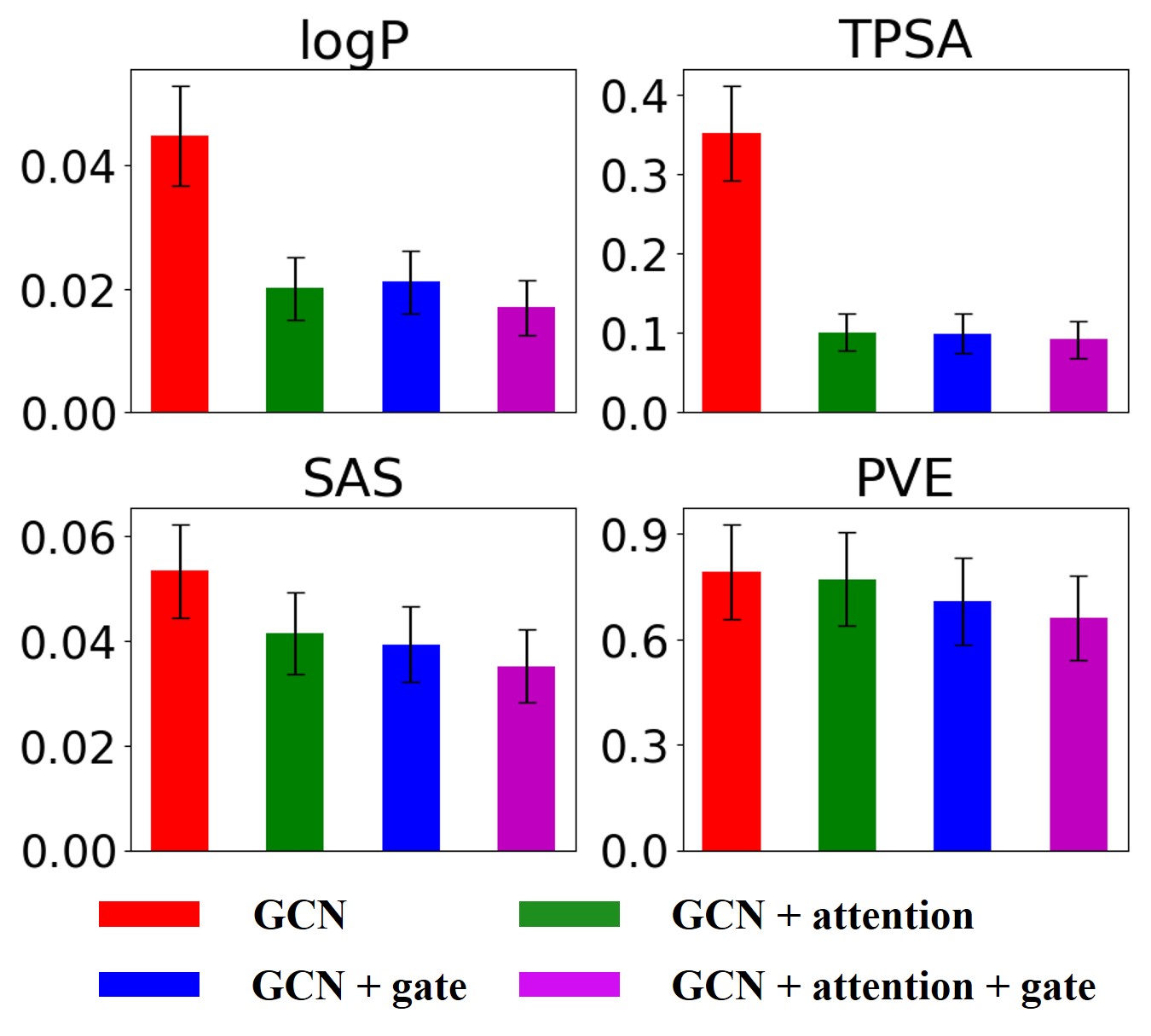}
 \caption{Performances of various GCN models for logP, TPSA, SAS, and PVE. We evaluate the performance by mean absolute errors (columns) and standard deviations of errors (error bars). Note that each error bar represents $0.1\times$standard deviation.}
 \label{fgr:fig4}
\end{figure}

Fig. 3 shows change in the mean absolute error (MAE) and total training time of partition coefficient (logP) prediction with respect to the number of graph convolution layers. The MAE of the vanilla GCN with a skip-connection began to increase as the number of graph convolution layers increased to three or more, which is consistent with the result of \citeauthor{kipf2016semi}. The independent use of attention mechanism and gated-skip connection attenuated such a trend up to five layers, but the MAE increases again after that. However, the simultaneous use of both the attention and gated skip-connection (GCN+attention+gate) resulted in a significant drop of the MAE even after the five layers. In addition, it achieved the smallest MAE among the best values of each model under the test, subsequently followed by the GCN+attention, the GCN+gate, and the vanilla GCN. In particular, the GCN+attention+gate showed comparable performance to that of GGNN. It should be noted that the former, however, is much faster than the latter, as shown in Fig. 3(b), because the latter employs a GRU for updating atom states.

\begin{figure*}[t!]
 \centering
 \includegraphics[width=18cm]{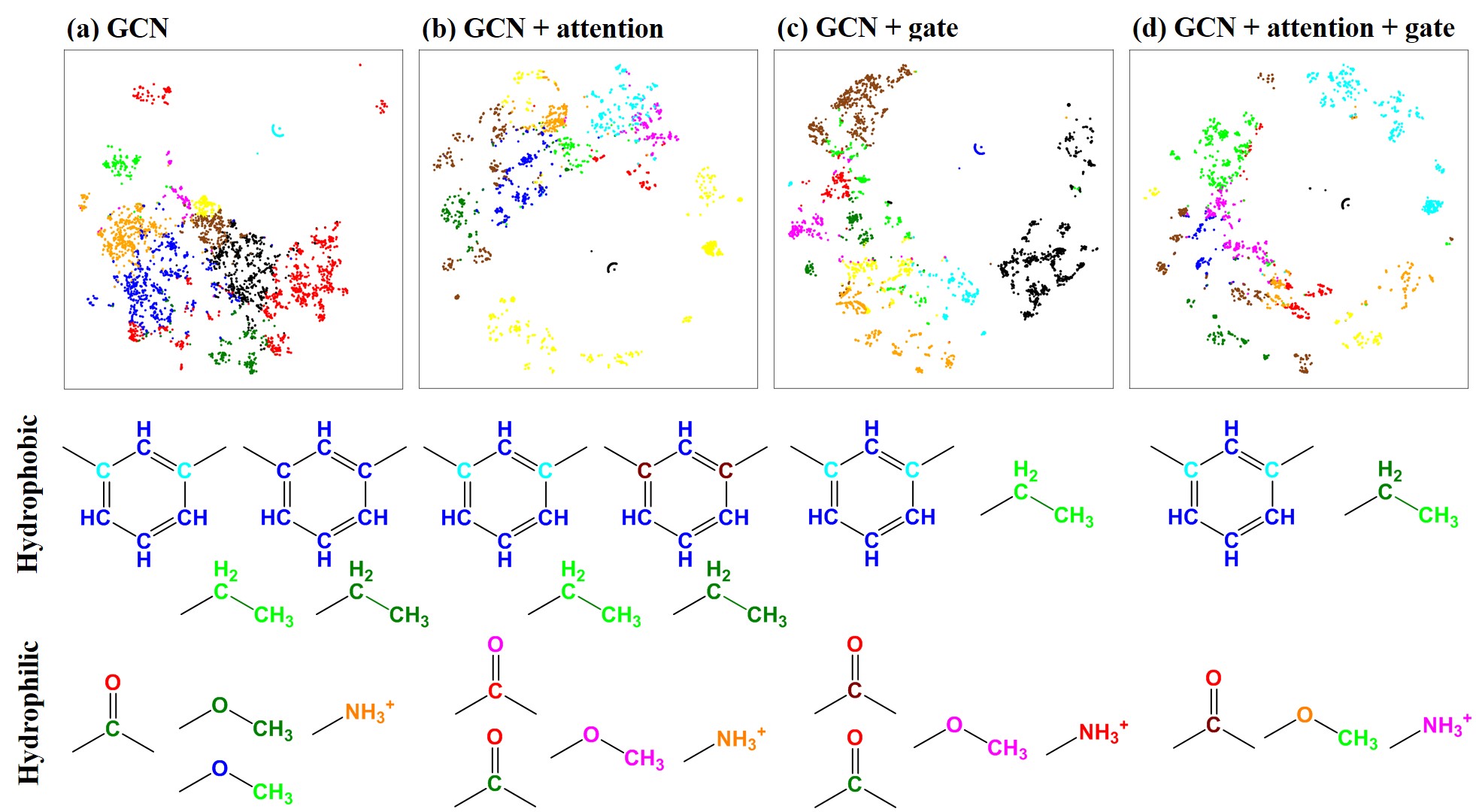}
 \caption{t-SNE visualization of atom features for logP extracted from (a) GCN, (b) GCN+attention, (c) GCN+gate, and (d) GCN+attention+gate (\textit{above}) and representative functional groups related to logP (\textit{below}). Each atom is colored with the atom labels obtained by k-means clustering of the atom features.}
 \label{fgr:fig5}
\end{figure*}

Fig. 4 shows more examples of the performance test for logP, topological surface area (TPSA), synthetic accessibility (SAS), and photovoltaic efficiency (PVE). We fixed the number of graph convolution layers to three and compared the results of the proposed models while keeping all other factors equally. Inclusion of the attention or the gated skip-connection significantly improved the performance of GCN, while the use of both functions showed the best performance. We note that the improvement effect on the PVE prediction was marginal compared to the other examples. This may be due to a relatively small number of training data points in the case of PVE (21,600) compared to the others (360,000) (see Implementation detail section for more details about the data sets).

\subsection{Interpretation of atom features for molecular structure-property relationships}

\begin{figure*}
 \centering
 \includegraphics[width=18cm]{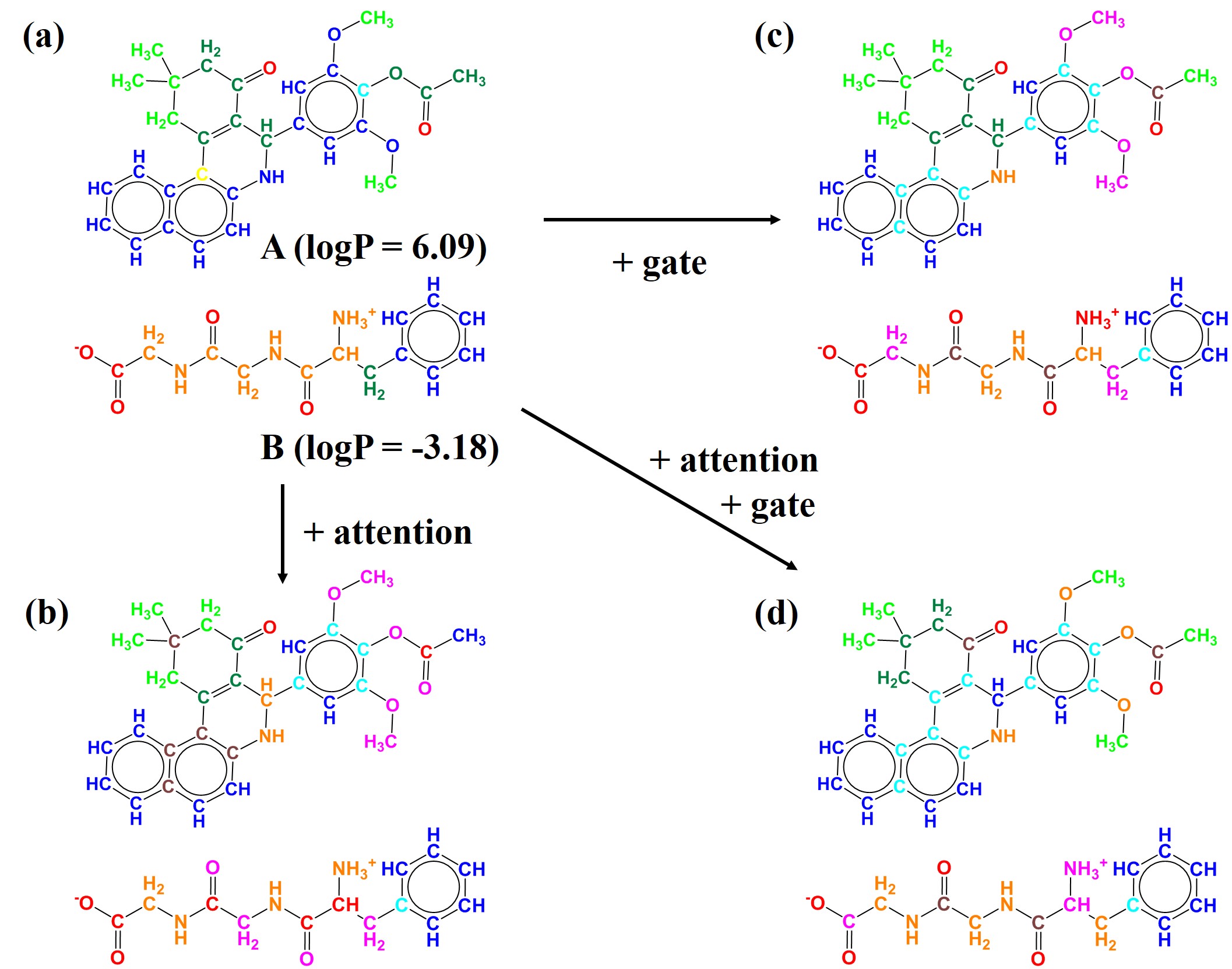}
 \caption{Representative molecules for log P. Each atom is colored with the atom labels obtained by k-means clustering. Atom features, which are used to determine the atom labels, are extracted from the (a) GCN, (b) GCN + attention, (c) GCN + gate, and (d) GCN + attention + gate. The molecule A has a large log P value (6.09), while the molecule B has a small log P value (-3.18).}
 \label{fgr:fig6}
\end{figure*}

\begin{figure*}
 \centering
 \includegraphics[width=18cm]{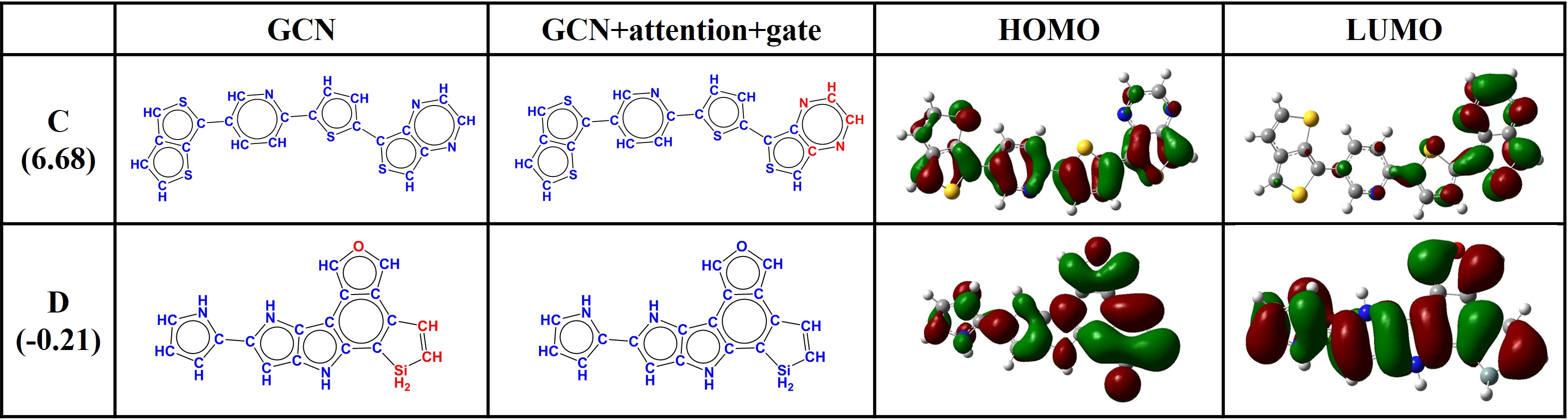}
 \caption{Two example molecules C and D with high and low PVE values, respectively. Each atom is colored with a binary color notation obtained from the same atom feature vector analysis with that in Fig. 6. The numbers in parentheses denote the true PVE values of the molecules.}
 \label{fgr:fig7}
\end{figure*}

DL is often called the feature learning or representation learning. As a result of correct mapping between inputs and outputs, DL models produce a high dimensional feature (or latent) space in which inputs with similar output values are closely located. In face recognition models, for example, convolution layers transform low level pixel values to high level features.\cite{krizhevsky2012imagenet} These high-level features represent the lips, ears, eyes and overall outline of the human face. In the case of natural language processing, words with similar meanings locate closely in a high dimensional feature space.\cite{mikolov2013efficient} In addition, vector operations such as `king - queen = man - woman' is performed among the feature vectors of words. In chemistry, similar sub-molecular structures (e.g., functional groups) would have similar atom features and hence contribute to a target property with similar extents. Thus, recognizing important substructures associated with the target property is critical to achieve a high accuracy. 
In this regard, we tried semantic interpretations of atom features for logP prediction as an example. We extracted atom feature vectors from the last graph convolution layer. Fig. 5 shows the 2D map of the high-dimensional latent space generated by the t-distributed stochastic neighbor embedding (t-SNE)\cite{maaten2008visualizing} of the atom feature vectors. Each color represents one of 10 groups classified by the k-means clustering\cite{arthur2007k}. In other words, all atoms have been classified into the 10 groups depending on their feature vectors associated with a given target property. We expected that atoms in a similar chemical environment have a same color because they would have similar feature vectors. For instance, we sampled a few functional groups significantly affecting logP values and labeled each atom with the colors obtained from the k-means clustering for each model. The results are displayed under each 2D map in Fig. 5. Overall, all models produced reasonable results. However, in the case of GCN, hydrophobic carbons have the same green color with hydrophilic carbons. Moreover, the same ether oxygen atoms have different colors. The GCN+attention and GCN+gate improved the results but do not fix all problems, e.g., the carbonyl carbons still have different colors. The GCN+attention+gate allowed more delicate classifications. Thus, some atoms in same functional groups share same colors across all molecules. 

To further investigate the dependence of atom features on local chemical environments, we analyzed two representative molecules for the logP prediction. We colored each atom of those molecules with the color notation obtained from each model in Fig. 5;  for example, the colors in Fig. 5(a) was used in Fig. 6(a). 
The vanilla GCN caused inconsistent color mapping. For instance, ether oxygens, aromatic carbons, and nitrogen have the same blue color for molecule A, despite that the aromatic carbons and the heteroatoms (oxygen and nitrogen) affect the log P value in opposite directions. In addition, carbonyl groups in molecule A (high logP value) have different colors with those in molecule B (low logP value), though they are in very similar chemical environments. Adopting the attention or the gated skip-connection improved the performance of the GCN by resolving some of such inconsistency. However, they still have limitations. Specifically, carbons connected to ether oxygens/nitrogen have the same pink/orange color with the oxygen/nitrogen atoms in Fig. 6(b). The simultaneous use of the attention and gated-skip connection resulted in more delicate classification. It could properly differentiate the heteroatoms from the neighboring carbon atoms as shown in Fig. 6(d). In addition, the carboxylic oxygens (red color in Fig. 6(d)) and the ether oxygens (orange color in Fig. 6(d)) were differentiated. Similar trends were observed across all molecules for logP, TPSA, and SAS as shown in Fig. S1, S2, and S3, respectively.

The logP values are directly related to molecular structures. Thus, it would be easy to identify specific functional groups determining the target value. Even undergraduate students majoring in chemistry would recognize those functional groups readily. However, identifying key molecular substructures related to the PVE would be challenging even for experts without any information on electronic structures. As the final example, we examined atom features of photovoltaic molecules obtained from the proposed models. In this case, we used only two colors (red and blue) to simplify the visual analysis. 

Fig. 7 shows the results of two representative molecules; molecule C has a very high PVE value, while molecule D has a very low PVE value. Interestingly, the GCN+attention+gate model categorized atoms of C into two groups with red and blue colors, while the GCN assigned only a single color to all atoms of the same molecule. On the contrary, the GCN assigned two colors to molecule D, whereas the GCN+attention+gate model assigned a single color. Importantly, we noted that the GCN+attention+gate model divided molecule C into two regions denoted by the two colors. To see whether these color mappings are closely related to key structural factors for the high PVE value or not, we plotted the highest occupied molecular orbital (HOMO) and the lowest unoccupied molecular orbital (LUMO) obtained from density functional theory calculations for molecules C and D (Fig. 7). Surprisingly, the red and blue regions of molecule C coincided with the areas to which the HOMO and LUMO are distributed. In the case of molecule D, all atoms had the same color, and indeed both the HOMO and the LUMO are delocalized over the entire region. The results can be rationalized as follows. To be a good photovoltaic molecule, an excited electron and the corresponding hole should be spatially separated to prevent an rapid recombination as well as to readily separate the electron-hole pair for energy harvesting. In this aspect, the augmented GCN was able to discover important structural features determining the PVE just from molecular graphs and raw input features shown in Fig. 1. In contrast, the atom feature mapping by the GCN showed opposite trends from those of the GCN+attention+gate model. The GCN poorly characterized atom features of photovoltaic materials. More examples can be found in Fig. S4. 

\begin{figure*}[t]
 \centering
 \includegraphics[width=18cm]{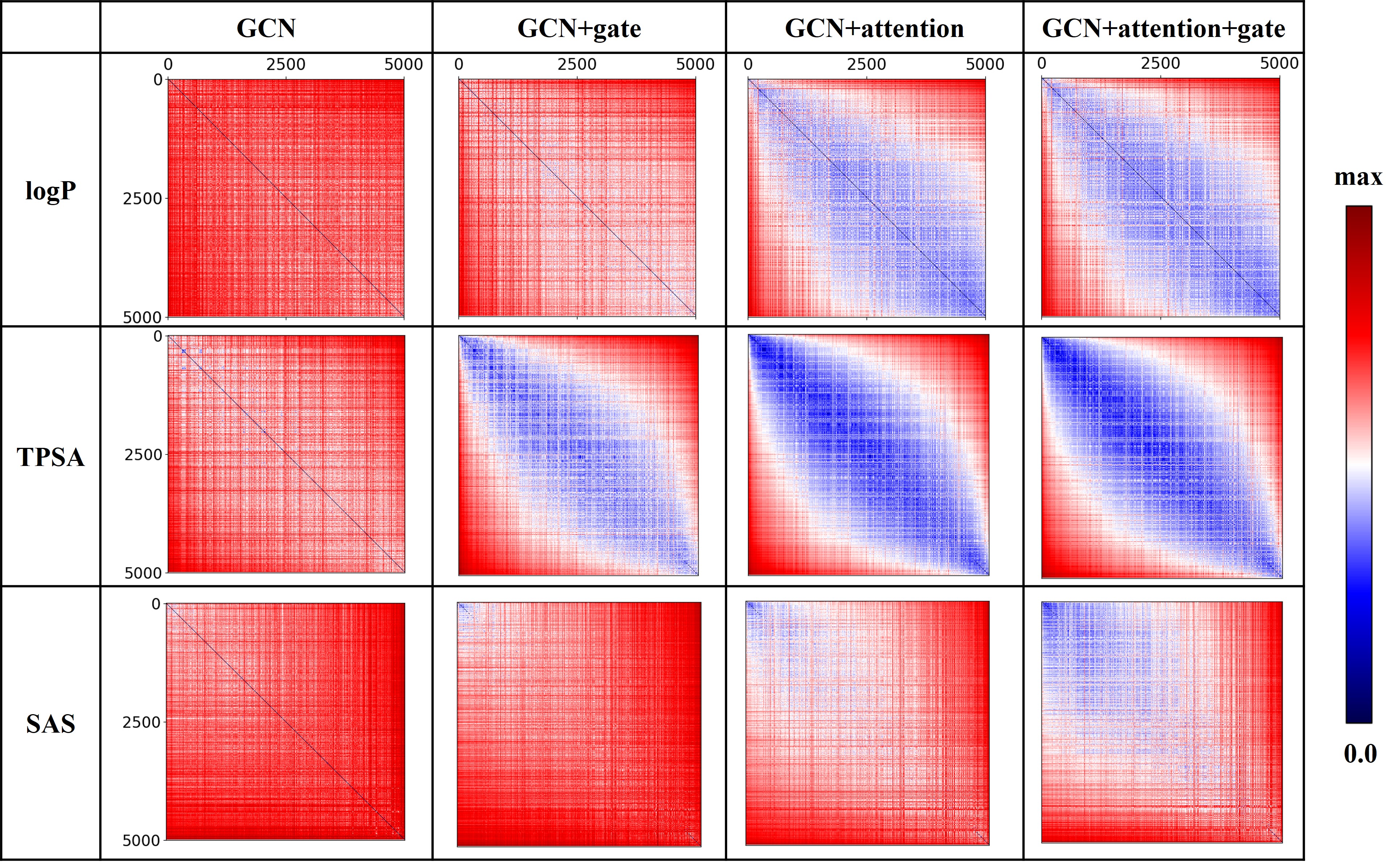}
 \caption{2D plot of the L2-norm distances between molecules in the latent space associated with logP, TPSA and SAS. 5,000 molecules chosen randomly are listed along the row and column in the ascending order of each molecular property. The distance between two molecules labeled by the row and column indices is denoted by the scale bar at the right side.}
 \label{fgr:fig8}
\end{figure*}

\subsection{Graph features and latent space}

The above examples manifest that the attention and gate mechanisms greatly help a graph convolution capture key structural features closely related to target properties. It is possible because adaptive attention weights allow the graph convolution to elaborately update individual atoms in different chemical environments. In addition, using the gate mechanism enables the graph convolution to update a present state by combining the previous state and an updated message state at an appropriate rate. As a result, the GCN+attention+gate model may produce a more desirable mapping between molecular structures and properties. Deeply learning the structure-property relationships is at the heart of successful molecular design. Previous works have shown how DL models utilize such relationships for \textit{de novo} molecular design. A common idea is to use a latent space which implies the structure-property mapping obtained from the DL models. In a well-trained latent space, molecules with similar properties are closely located to each other. Thus, we can generate new molecules with desired properties by exploring the latent space. 

Along the same lines, we examined the latent space for the four properties. We used the latent vector, $z_G$,  obtained from the readout layer for each property as shown in Fig. 2(c) and (d). To statistically analyze the results, we sorted 5,000 molecules randomly chosen from each test set in an ascending order of their properties and measured the L2-norm distance ($d_{ij} = \norm{ \mathbf{z_{G_i}} - \mathbf{z_{G_j}} }_2 $) between them. Fig. 8 shows the 2D plot of the normalized distance mapping. The distance between two molecules labeled by row and column indices is denoted by the scale bar on the right side. The diagonal elements colored in dark blue indicate the distance of a molecule from itself, while the dark red elements denote the farthest distance. It seems that the GCN does not show any meaningful patterns in the distance mapping. Even though molecules have similar logP values, they are not closely located in the latent space. However, adopting the attention or gated skip-connection puts molecules with similar properties closer to each other in the space - a tendency more conspicuous when using the attention.  The GCN+attention+gate model  shows a gradual color change from dark blue to dark red, resulting in unique patterns. Such a gradual color change hints that molecules with a similar property tend to be clustered in the latent space. These results manifest that the augmented GCN models especially with both attention and gated skip-connection learn the structure-property relationship more deeply than the vanilla GCN does.

\section{Conclusion}
We have shown that the attention- and gate-augmented graph convolutional network (GCN) model outperforms the vanilla GCN in supervised learning of various molecular properties. The attention mechanism can identify atoms in different chemical environments by considering the importance of their neighbor atoms. The gated skip-connection is used for the update of atom states at a certain rate. Thus, the augmented GCN can extract important structural features which better determine a target property. By analyzing atom features, we demonstrated that the augmented GCN was able to identify polar and nonpolar functional groups of molecules as key structural features for logP. More interestingly, it identified two distinct parts of molecules as important structural features for high photovoltaic efficiency. The two distinct parts correspond to donor and acceptor orbitals of the molecules in charge-transfer excitations. Evidently the augmented GCN elucidates the structure-property relationship from chemical data better than the GCN does. As a result, it produced well-trained latent spaces where molecules with similar properties were closely located to each other. 

Such a high performance of the improved GCN offers various application possibilities. In particular, accurate learning of the structure-property relationship is essential to design new molecules with desired properties by using molecular generative models. For instance, \citeauthor{gomez2018automatic} showed that a variational autoencoder can generate new molecules with a target property through the gradient-based optimization process in a latent space.\cite{gomez2018automatic} We also demonstrated that a conditional variational autoencoder can design molecules with simultaneous control of multiple target molecular properties for drug discovery by embedding them directly in latent vectors.\cite{lim2018molecular} \citeauthor{segler2017generating} and \citeauthor{gupta2018generative} designed molecules with specific biological activities using a natural language processing model combined with transfer learning.\cite{segler2017generating, gupta2018generative} \citeauthor{jaques2016sequence}, \citeauthor{olivecrona2017molecular}, and \citeauthor{guimaraes2017objective} proposed methods which finely tune a pretrained generative model using reinforcement learning to generate molecules with certain desirable properties.\cite{jaques2016sequence, olivecrona2017molecular, guimaraes2017objective}

However, these models utilized the SMILES representation of molecular structures, so they may have a low rate of valid molecules due to lack of topological information. Recently, interest in developing graph-based generative models has been growing.\cite{li2018learning, you2018graphrnn, jin2018junction, simonovsky2018graphvae, li2018multi} The early stage studies showed promising results with high rates of valid and novel molecules, which is important to exploring an extended region of chemical space. In many studies, GGNN is used to update atom states. We echo that the augmented GCN shows a comparable accuracy to GGNN but with much smaller training and test time. We expect that it will substantially improve such generative models for \textit{de novo} molecular design via a well-trained structure-property relationship. Consequently, we believe that our proposal and interpretation have a broad impact and would give insights on successful molecular engineering.

\section*{Implementation detail}
\subsection*{Setup for training, validation, and test}
Table 2 summarizes configurations of the deep learning models proposed in this work. We randomly split total 500,000  molecules in ZINC dataset\cite{irwin2005zinc} and 29,978 in Havard Clean Energy Project dataset\cite{hachmann2011harvard} into a ratio of [0.72 : 0.18 : 0.1] to carry out [training : validation : test].

\begin{table}[h]
\small
  \label{tbl:table2}
    \caption{Configurations of the models in this work}
  \begin{tabular*}{0.5\textwidth}{@{\extracolsep{\fill}}llll}
    \hline
     & logP, TPSA, SAS & PVE  \\
    \hline
    Dataset & ZINC & Clean Energy Project \\
    Number of training data & 360,000 & 21,600 \\
    Number of validation data & 90,000 & 5,400 \\
    Number of test data & 50,000 & 2,978 \\
    Batch size & 100 & 100 \\
    Number of epochs & 100 & 200 \\
    Optimizer & Adam & Adam \\
    Learning rate & 0.001 & 0.001 \\
    Decay rate & 0.95 & 0.97
     \\
    \hline
  \end{tabular*}
\end{table}

Output dimensions of all graph convolution layers are 32 and fully-connected layers for predictor are 512.  For example, the output dimensions for all hidden layers in GCN (and GCN+attention+gate) with three graph convolution layers are [32 - 32 - 32 - 512 - 512 - 512 - 1]. We used TensorFlow TensorFlow\cite{abadi2016tensorflow} for implementations. We uploaded our codes on S. Ryu's GitHub(\url{http://github.com/seongokryu/augmented-gcn}). 

\section*{Acknowledgements}
This work was supported by Basic Science Research Programs through the National Research Foundation of Korea (NRF) funded by the Ministry of Science, ICT and Future Planning (NRF-2017R1E1A1A01078109).

\section*{Author contributions}
S.R. and W.Y.K. conceived the idea, S.R. did the implementation and run the simulation. All the authors analyzed the results and wrote the manuscript together.

\section*{Conflicts of interest}
The authors declare no competing financial interests.
\bibliography{rsc} 
\bibliographystyle{rsc} 

\end{document}